\documentclass[10pt,twocolumn,letterpaper]{article}

\usepackage{cvpr}
\usepackage{times}
\usepackage{epsfig}
\usepackage{graphicx}
\usepackage{amsmath}
\usepackage{amssymb}

\usepackage{booktabs}
\usepackage{subcaption}
\newcommand\norm[1]{\left\lVert#1\right\rVert}
\usepackage{array}
\newcolumntype{P}[1]{>{\centering\arraybackslash}p{#1}}

\DeclareSymbolFont{extrasymbols}{OMS}{cmsy}{m}{n}
\DeclareMathDelimiter{\lVert}{\mathopen}{extrasymbols}{"6B}{largesymbols}{"0D}
\DeclareMathDelimiter{\rVert}{\mathclose}{extrasymbols}{"6B}{largesymbols}{"0D}

\usepackage[breaklinks=true,bookmarks=false]{hyperref}

\cvprfinalcopy 

\def\cvprPaperID{9208} 

\begin{document}

\title{Improving the Robustness of Capsule Networks to Image Affine Transformations}

\author{
Jindong Gu \\
University of Munich\\
Siemens AG, Corporate Technology\\
{\tt\small  jindong.gu@siemens.com} \\
\and
Volker Tresp \\
University of Munich\\
Siemens AG, Corporate Technology\\
{\tt\small volker.tresp@siemens.com} \\
}

\maketitle

\begin{abstract}
Convolutional neural networks (CNNs) achieve translational invariance by using pooling operations. However, the operations do not preserve the spatial relationships in the learned representations. Hence, CNNs cannot extrapolate to various geometric transformations of inputs. Recently, Capsule Networks (CapsNets) have been proposed to tackle this problem. In CapsNets, each entity is represented by a vector and routed to high-level entity representations by a dynamic routing algorithm. CapsNets have been shown to be more robust than CNNs to affine transformations of inputs. However, there is still a huge gap between their performance on transformed inputs compared to untransformed versions. In this work, we first revisit the routing procedure by (un)rolling its forward and backward passes. Our investigation reveals that the routing procedure contributes neither to the generalization ability nor to the affine robustness of the CapsNets. Furthermore, we explore the limitations of capsule transformations and propose affine CapsNets (Aff-CapsNets), which are more robust to affine transformations. On our benchmark task, where models are trained on the MNIST dataset and tested on the AffNIST dataset, our Aff-CapsNets improve the benchmark performance by a large margin (from 79\% to 93.21\%), without using any routing mechanism.
\end{abstract}

\section{Introduction}
Human visual recognition is quite insensitive to affine transformations. For example, entities in an image, and a rotated version of the entities in the image, can both be recognized by the human visual system, as long as the rotation is not too large. Convolutional Neural Networks (CNNs), the currently leading approach to image analysis,  achieve affine robustness by training on a large amount of data that contain different transformations of target objects. Given limited training data, a common issue in many real-world tasks, the robustness of CNNs to novel affine transformations is limited \cite{sabour2017dynamic}.

With the goal of learning image features that are more aligned with human perception, Capsule Networks (CapsNets) have recently been proposed \cite{sabour2017dynamic}. The proposed CapsNets differ from CNNs mainly in two aspects: first, they represent each entity by an activation vector, the magnitude of which represents the probability of its existence in the image; second, they assign low-level entity representations to high-level ones using an iterative routing mechanism (a dynamic routing procedure). Hereby, CapsNets aim to keep two important features: equivariance of output-pose vectors and invariance of output activations. The general assumption is that the disentanglement of variation factors makes CapsNets more robust than CNNs to affine transformations.

The currently used benchmark task to evaluate the affine robustness of a model is to train the model on the standard MNIST dataset and test it on the AffNIST\footnote{Each example is an MNIST digit with a small affine transformation.} dataset. CapsNets achieve 79\% accuracy on AffNIST, while CNNs with similar network size only achieve 66\% \cite{sabour2017dynamic}. Although CapsNets have demonstrated their superiority on this task, there is still a huge performance gap since CapsNets achieve more than 99\% on the untransformed MNIST test dataset.

In our paper, we first investigate the effectiveness of components that make CapsNets robust to input affine transformations, with a focus on the routing algorithm. Many heuristic routing algorithms have been proposed \cite{hinton2018matrix,wang2018optimization,lenssen2018group} since \cite{sabour2017dynamic} was published. However, recent work \cite{Paik2019CapsuleNN} shows that all routing algorithms proposed so far perform even worse than a uniform/random routing procedure.

From both numerical analysis and empirical experiments, our investigation reveals that the dynamic routing procedure contributes neither to the generalization ability nor to the affine robustness of CapsNets. Therefore, it is infeasible to improve the affine robustness by modifying the routing procedure. Instead, we investigate the limitations of the CapsNet architectures and propose a simple solution. Namely, we propose to apply an identical transformation function for all primary capsules and replace the routing by a simple averaging procedure (noted as No Routing).

Our contributions of this work can be summarized as follows: 1) We revisit the dynamic routing procedure of CapsNets; 2) We investigate the limitations of the current CapsNet architecture and propose a more robust affine Capsule Networks (Aff-CapsNet); 3) Based on extensive experiments, we investigate the properties of CapsNets trained without routing. Besides, we demonstrate the superiority of Aff-CapsNet.

The rest of this paper is organized as follows: Section \ref{sec:related} first reviews CapsNets and related work. Section \ref{sec:effect} investigates the effectiveness of the routing procedure by (un)rolling the forward and backward passes of the iterative routing iterations. Section \ref{sec:aff} shows the limitations of current CapsNets on the affine transformations and proposes a robust affine CapsNet (Aff-CapsNet). Section \ref{sec:aff} conducts extensive experiments to verify our findings and proposed modifications. The last two sections discuss and conclude our work.

\section{Background and Related Work}
\label{sec:related}
In this section, we first describe the CapsNets with dynamic routing and then review related work.

\subsection{Fundamentals of Capsule Networks}
CapsNets \cite{sabour2017dynamic} encode entities with capsules. Each capsule is represented by an activity vector (e.g., the activation of a group of neurons), and elements of each vector encode the properties of the corresponding entity. The length of the activation vector indicates the confidence of the entity's existence. The output classes are represented as high-level capsules.

A CapsNet first maps the raw input features to low-level capsules and then routes the low-level capsules to high-level ones. For instance, in image classification tasks, a CapsNet starts with one (or more) convolutional layer(s) that convert the pixel intensities into low-level visual entities. A following capsule layer of the CapsNet routs low-level visual entities to high-level visual entities. A CapsNet can have one or more capsule layers with routing procedures.

Given a low-level capsule $\pmb{u}_i$ of the $L$-th layer with $N$ capsules, a high-level capsule $\pmb{s}_j$ of the $(L+1)$-th layer with $M$ capsules, and a transformation matrix $\pmb{W}_{ij}$, the routing process is
\begin{equation}
\pmb{\hat{u}}_{j|i}=\pmb{W}_{ij} \pmb{u}_i, \qquad  \pmb{s}_j=\sum^N_i c_{ij} \pmb{\hat{u}}_{j|i}
\label{equ:trans}
\end{equation}
where $c_{ij}$ is a coupling coefficient that models the degree with which $\pmb{\hat{u}}_{j|i}$ is able to predict $\pmb{s}_j$. The capsule $\pmb{s}_j$ is shrunk to a length in (0, 1) by a non-linear squashing function $g(\cdot)$, which is defined as
\begin{equation}
\pmb{v}_j= g(\pmb{s}_j) = \frac{\norm{\pmb{s}_j}^2}{1+\norm{\pmb{s}_j}^2} \frac{\pmb{s}_j}{\norm{\pmb{s}_j}}
\label{equ:squa}
\end{equation}

The coupling coefficients $\{c_{ij}\}$ are computed by an iterative routing procedure. They are updated so that high agreement ($a_{ij} =  \pmb{v}^T_j \pmb{\hat{u}}_{j|i}$) corresponds to a high value of $c_{ij}$. 
\begin{equation}
c_{ij}= \frac{\exp(b_{ij})}{\sum_k \exp(b_{ik})}
\label{equ:coup}
\end{equation}
where initial logits $b_{ik}$ are the log prior probabilities and updated with $b_{ik} = b_{ik} + a_{ij} $ in each routing iteration. The coupling coefficients between a $i$-th capsule of the $L$-th layer and all capsules of the $(L+1)$-th layer sum to  1, i.e., $\sum_{j=1}^M c_{ij} =1$. The steps in Equations \ref{equ:trans}, \ref{equ:squa}, and \ref{equ:coup} are repeated $K$ times in the routing process, where $\pmb{s}_j$ and $c_{ij}$ depend on each other. 

\subsection{Related Work}
\textbf{Routing Algorithms:} Many papers have improved the routing-by-agreement algorithm. \cite{zhang2018fast} generalizes existing routing methods within the framework of weighted kernel density estimation and proposes two fast routing methods with different optimization strategies. \cite{Choi2019AttnCaps} proposes an attention-based routing procedure with an attention module, which only requires a fast forward-pass. The agreement $a_{ij}$ can also be calculated based on a Gaussian distribution assumption \cite{hinton2018matrix,bahadori2018spectral} or distance measures \cite{lenssen2018group} instead of the simple inner product.

Since the routing procedure is computationally expensive, several works propose solutions reducing the complexity of the iterative routing process. \cite{wang2018optimization} formulates the routing strategy as an optimization problem that minimizes a combination of clustering-like loss and a KL distance between the current coupling distribution and its last states. \cite{li2018neural} approximates the expensive routing process with two branches: a master branch that collects primary information from its direct contact in the lower layer and an aide branch that replenishes the master branch based on pattern variants encoded in other lower capsules.

\textbf{Understanding the Routing Procedure:} \cite{chen2018generalized} incorporates the routing procedure into the training process by making coupling coefficients trainable, which are supposed to be determined by an iterative routing process. The coupling coefficients are independent of examples, which stay unchanged in the testing phase. What they proposed is simply to reduce the iterative updates to a single forward pass with prior coupling coefficients. \cite{anonymous2020capsule} removes the routing procedure completely and modifies the CapsNet architectures. Their pure CapsNets achieve competitive performance. However, it has not been investigated how the properties of their CapsNets, e.g., the robustness to affine transformation, will be affected by the removal of the routing procedure. Furthermore, \cite{Paik2019CapsuleNN} shows that many routing procedures \cite{sabour2017dynamic,hinton2018matrix,wang2018optimization,lenssen2018group} are heuristic, and perform even worse than a random routing assignment.

\section{Revisiting the Dynamic Routing of CapsNets}
\label{sec:effect}
In this section, we analyze dynamic routing, both theoretically and empirically. By unrolling the backpropagation of the routing procedure and rolling the forward propagation of the routing procedure, we show which role the routing procedure plays in CapsNets.

\begin{figure}
  \includegraphics[width=\linewidth]{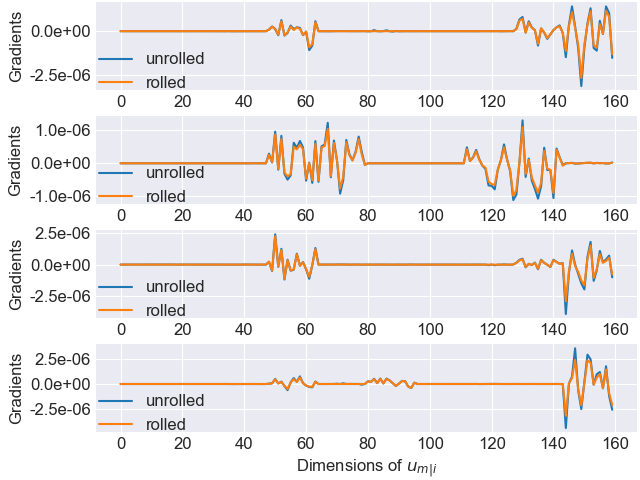}
  \caption{The gradients of the loss w.r.t. randomly choosen $\pmb{\hat{u}}_{m|i}$ are visualized. The blue lines correspond to the unrolled routing iterations in Gradient Backpropagation, while the yellow lines to rolled routing iterations.}
  \label{fig:grad}
\end{figure}

\subsection{Backpropagation through Routing Iterations}
\label{sec:rolled}
The forward pass of an iterative routing process can be written as the following iterative steps
\begin{equation}
\small
\begin{split}
\pmb{s}^{(t)}_j &= \sum^N_i c^{(t)}_{ij} \pmb{\hat{u}}_{j|i}  \\
\pmb{v}^{(t)}_j &= g(\pmb{s}^{(t)}_j)  \\
c^{(t+1)}_{ij}     &= \frac{ \exp(b_{ij} + \sum_{r=1}^t \pmb{v}^{(r)}_j \pmb{\hat{u}}_{j|i}) }{\sum_k  \exp(b_{ik} + \sum_{r=1}^t \pmb{v}^{(r)}_k \pmb{\hat{u}}_{k|i} )}
\end{split}
\end{equation}
where the superscript $t \in \{1, 2, ...\}$ is the index of an iteration. The $c^{(1)}_{ij}$ and $b_{ij}$ are initialized as in Equation \ref{equ:coup}. 

Assuming that there are $K$ iterations and the classification loss is $\mathcal{L}(\pmb{y}, \pmb{t})$, where $\pmb{y}=(\lVert\pmb{v}^{(K)}_1\rVert, \cdots, \lVert\pmb{v}^{(K)}_M\rVert)$ is the prediction and $\pmb{t}$ the target, the gradients through the routing procedure are
\begin{equation}
\small
\frac{\partial \mathcal{L}}{\partial \pmb{\hat{u}}_{m|i}} = \frac{\partial \mathcal{L}}{\partial \pmb{v}^{(K)}_{m}} \frac{\partial \pmb{v}^{(K)}_{m}}{\partial \pmb{s}^{(K)}_m} c^{(K)}_{im} + \sum^{M}_{j=1} \frac{\partial \mathcal{L}}{\partial \pmb{v}^{(K)}_{j}} \frac{\partial \pmb{v}^{(K)}_{j}}{\partial \pmb{s}^{(K)}_j} \pmb{\hat{u}}_{j|i} \frac{\partial c^{(K)}_{ij}}{\partial \pmb{\hat{u}}_{m|i}}
\label{equ:back}
\end{equation}
The gradients are propagated through the unrolled routing iteration via the second item of Equation \ref{equ:back}, which is also the main computational burden of the expensive routing procedure in CapsNets. By unrolling this term,  we prove that 
\begin{equation}
\small
\frac{\partial \mathcal{L}}{\partial \pmb{\hat{u}}_{m|i}} \approx C \cdot \frac{\partial \mathcal{L}}{\partial \pmb{v}^{(K)}_{m}} \frac{\partial \pmb{v}^{(K)}_{m}}{\partial \pmb{s}^{(K)}_m} c^{(K)}_{im}
\label{equ:cond}
\end{equation}
where $C$ is a constant, which can be integrated into the learning rate in the optimization process (see the proof in Appendix A). The approximation means that the gradients flowing through $c^{(K)}_{ij}$ in Equation \ref{equ:back} can be ignored. The $c^{(K)}_{ij}$ can be treated as a constant in Gradient Backpropagation, and the routing procedure can be detached from the computational graph of CapsNets.

To confirm Equation \ref{equ:cond} empirically, we visualize $\frac{\partial \mathcal{L}}{\partial \pmb{\hat{u}}_{m|i}}$. Following \cite{sabour2017dynamic}, we train a CapsNet on the MNIST dataset. The architecture and the hyper-parameter values can be found in Appendix B. We first select capsule predictions $\pmb{\hat{u}}_{j|i}$ randomly prior to the routing process and then visualize their received gradients in two cases: 1) unrolling the routing iterations as in \cite{sabour2017dynamic}; 2) rolling the routing iterations by taking all $c_{ij}$ as constants in Gradient Backpropagation (i.e., ignoring the second item in Equation \ref{equ:back}). As shown in each plot of Figure \ref{fig:grad}, the gradients of the two cases (blue lines and yellow lines) are similar to each other.

In this section, we aim to show that the intrinsic contribution of the routing procedure is to identify specified constants as coupling coefficients $c^{(K)}_{ij}$. Without a doubt, both computational cost and memory footprint can be saved by rolling the routing iterations in Gradient Backpropagation. The computational graphs of the two cases can be found in Appendix C.

\subsection{Forward Pass through Routing Iterations}
The forward iterative routing procedure can be formulated as a function, mapping capsule predictions $\pmb{\hat{u}}$ to coupling coefficients, i.e., $ \pmb{\hat{u}} \rightarrow \pmb{C}^{(K)} = \{c^{(K)}_{ij}\}$ where the indexes of low-level capsules $i$ vary from $1$ to $N$ and the indexes of high-level capsules $j$ vary from $1$ to $M$. Given an instance, without loss of generality, we assume the ground-truth class is the $M$-th (i.e., $\pmb{v}_{M}$). With the idea behind the CapsNet, the optimal coupling coefficients $\pmb{C}^* = \{c^*_{ij}\}$ of the instance can be described as
\begin{equation}
\small
\begin{split}
\pmb{C}^* =\max_{\{c_{ij}\}} f(\pmb{\hat{u}}) & = \max_{\{c_{ij}\}}  (\sum^N_i c_{iM}  \pmb{\hat{u}}_{M|i} g(\sum_i c_{iM} \pmb{\hat{u}}_{M|i} ) \\ & -  \sum^{M-1}_j \sum^N_i c_{ij}  \pmb{\hat{u}}_{j|i} g(\sum_i c_{ij} \pmb{\hat{u}}_{j|i} ) )
\end{split}
\end{equation}
where the first term describes the agreement on the target class, and the second term corresponds to the agreement on non-ground-truth classes. The optimal coupling coefficient $\pmb{C}^*$ corresponds to the case where the agreement on the target class is maximized, and the agreement on the non-ground-truth classes is minimized.

\begin{figure}
  \includegraphics[width=0.93\linewidth]{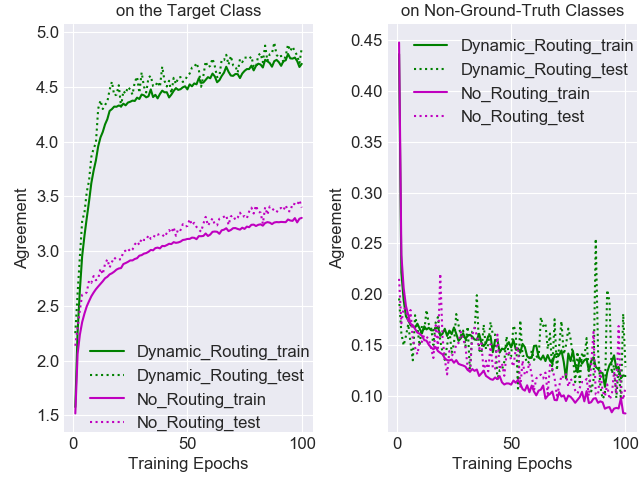}
  \caption{The green lines correspond to the model with dynamic routing, while the magenta ones to the model without routing procedure. For both models, the agreement on the target class increases with training time, and the agreement on the non-ground-truth classes decreases. The values are averaged over the whole training or test dataset.}
\label{fig:cc_implicit}
\end{figure}

Many routing algorithms differ only in how they approximate $\pmb{C}^*$. For instance, the original work \cite{sabour2017dynamic} approximates $\pmb{C}^*$ with an iterative routing procedure. Without requiring iterative routing steps, \cite{chen2018generalized} makes $\{b_{ij}\}$ trainable to approximate $\{c^*_{ij}\}$. Their proposal can be understood as only one-step routing with learned prior coupling coefficients. By further reformulation, we show that the optimal $\pmb{s}^*_j$ can be learned, without a need for coupling coefficients, as
\begin{equation}
\small
\pmb{s}^*_j = \sum^N_i c^*_{ij} \pmb{\hat{u}}_{j|i} =  \sum^N_i c^*_{ij} \pmb{W}^*_{ij} \pmb{u}_i =  \sum^N_i \pmb{W}'_{ij} \pmb{u}_i
.\end{equation}
In the training process, the transformation matrix $\pmb{W}_{ij}$ is updated via Gradient Decent Method. The coupling coefficients $c_{ij}$ are determined by the agreement between low-level capsules and the corresponding high-level capsules. The training process ends up with parameter values $\pmb{s}^*_j, \pmb{W}^*_{ij}, c^*_{ij}$. As shown in Equation 8, the CapsNet can achieve the same results by simply learning a transformation matrix $\pmb{W}'_{ij}$ without $c^*_{ij}$. In other words, the connection strengths $c^*_{ij}$ between low-level capsules and high-level capsules can be learned implicitly in the transformation matrix $\pmb{W}'_{ij}$. Therefore, we can conclude that different ways to approximate $\pmb{C}^*$ do not make a significant difference since the coupling coefficients will be learned implicitly.

We visualize the implicit learning process of the coupling coefficients. In our experiments, we introduce the no-routing approach, where we remove the iterative routing procedure by setting all coupling coefficient  $c_{ij}$ as a constant $\frac{1}{M}$. In each training epoch, the agreement on the target class and on the non-ground-truth classes is visualized in Figure \ref{fig:cc_implicit}. As a comparison, we also visualize the corresponding agreement values of CapNets with the dynamic routing process. We can observe that, during the training process, the agreement on the target class increases (in the left plot) for both cases, and the agreement on the non-ground-truth classes decreases (in the right plot). In other words, $f(\pmb{\hat{u}})$ increases in both CapNets with/without routing procedure, meaning that the coupling coefficients can be learned implicitly.

In summary, the affine robustness of CapsNet can not be contributed to the routing procedure. We conclude that it is not infeasible to improve the robustness of CapsNet by modifying the current routing-by-agreement algorithm.

\begin{figure*}
\centering
  \includegraphics[width=0.98\linewidth]{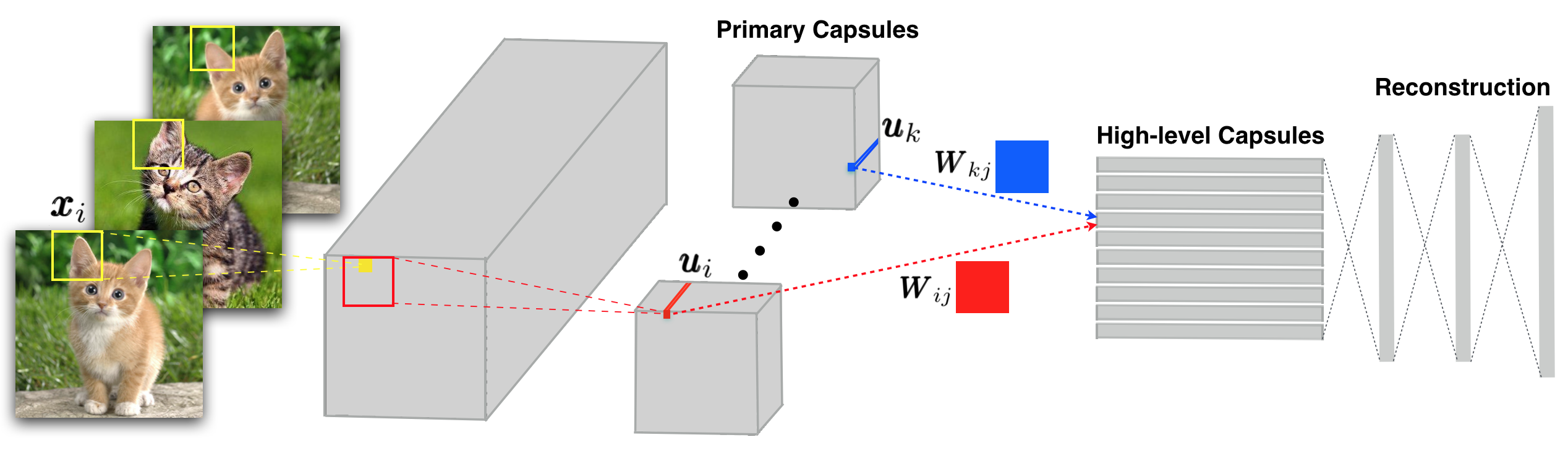}
  \caption{Illustration of the limitations of CapsNets: The transformation matrix $\pmb{W}_{ij}$ can only transform $\pmb{u}_{i}$ to high-level capsules, while $\pmb{W}_{kj}$ can only make meaningful transformations on $\pmb{u}_{k}$.  When an input is transformed (e.g., rotated), the receptive field corresponding to $\pmb{u}_{i}$ is not $\pmb{x}_{i}$ any more. For the novel $\pmb{u}_{i}$, the transformation process using $\pmb{W}_{ij}$ can fail.}
  \label{fig:overview}
\end{figure*}

\section{Affine Robustness of Capsule Networks}
\label{sec:aff}
Besides the dynamic routing process, the other difference between CapsNets and traditional CNNs is the CapsNet architecture. CapsNets represent each entity with a capsule and transform it to high-level entities employing transformation matrices. In this section, we investigate the limitation of the transformation process in terms of affine robustness and propose robust affine capsule networks.

\subsection{The Limitation of CapsNets} The CapsNet starts with two convolutional layers, which converts the pixel intensities to form primary (low-level) capsules (e.g., the red cuboid in Figure \ref{fig:overview} is a capsule $\pmb{u}_i$). Each primary capsule has a certain receptive field (e.g., the image patch $\pmb{x}_i$ marked with the yellow rectangle). For all inputs, the coordinates of the receptive field of $\pmb{u}_i$ are the same. In other words, a primary capsule can only see a specific area in input images. We denote the corresponding converting process by $\pmb{u}_i = p_i(\pmb{x}_i)$.

Each primary capsule is transformed to high-level capsules with the corresponding transformation matrix. Each transformation matrix $\pmb{W}_{ij}$ learns how to transform the $i$-th low-level capsule to the $j$-th high-level one, i.e., $\pmb{\hat{u}}_{j|i} = t_{j|i}(\pmb{u}_i)$. The transformation process corresponding to the input patch $\pmb{x}_i$ can be described as 
\begin{equation}
\small
\pmb{\hat{u}}_{j|i} = \pmb{W}_{ij} \pmb{u}_i = t_{j|i}(\pmb{u}_i) = t_{j|i}(p_i(\pmb{x}_i))
.\end{equation}
The transformation matrix $\pmb{W}_{ij}$ can only make meaningful transformations for the entities that have, at some point, appeared in the position of $\pmb{x}_i$. The input domain of the transformation function $t_{j|i}(\cdot)$ is $\mathbb{U}_i$.

In the testing phase, if novel affine transformations are conducted on the input, the corresponding transformation process $t_{j|i}(p_i(\pmb{x}'_i))$ are not meaningful since $p_i(\pmb{x}'_i)$ is not in the input domain $\mathbb{U}_i$. In other words, the transformation matrix $\pmb{W}_{ij}$ does not describe a meaningful transformation since the entities of $\pmb{x}'_i$ have never appeared in the position of the patch $\pmb{x}_i$ during training. Hence, the CapsNet is limited in its generalization ability to novel affine transformations of inputs.

\subsection{Robust Affine Capsule Networks} To overcome the limitation above, we propose a very simple but efficient solution. Concretely, we propose to use the same transformation function for all primary capsules (i.e., ensuring $t_{j|i}(\cdot) \equiv t_{j|k}(\cdot)$). We implement a robust affine capsule network (Aff-CapsNet) by sharing a transformation matrix. Formally, for Aff-CapsNets, we have
\begin{equation}
\small
\pmb{W}_{ij} = \pmb{W}_{kj}, \; \, \forall \, i, k \in \{1, 2, \cdots, N\}
\end{equation}
where $N$ is the number of primary capsules. In Aff-CapsNets, the transformation matrix can make a meaningful transformation for all primary capsules since it learns how to transform all low-level capsules to high-level capsules during training. The transformation matrix sharing has also been explored in a previous publication \cite{rajasegaran2019deepcaps}. The difference is that they aim to save parameters, while our goal is to make CapsNets more robust to affine transformations.

From another perspective, primary capsules and high-level capsules correspond to local coordinate systems and global ones, respectively. A transformation matrix is supposed to map a local coordinate system to the global one. One might be wondering that the transformation from each local coordinate system to a global one requires a specific transformation matrix. In existing architectures, the coordinate system is high-dimensional. Hence, a single shared transformation matrix is able to make successful transformations for all local coordinate systems.

\section{Experiments and Analysis}
The experiments include two parts: 1) We train CapsNets with different routing mechanisms (including no routing) on popular standard datasets and compare their properties from many perspectives; 2) We show that Aff-CapsNets outperform CapsNets on the benchmark dataset and achieves state-of-the-art performance. For all the experiments of this section, we train models with 5 random seeds and report their averages and variances.

\begin{table*}
\begin{center}
\begin{tabular}{rP{1.8cm}P{1.8cm}P{1.8cm}P{1.8cm}}
\toprule
Datasets & MNIST & FMNIST & SVHN & CIFAR10  \\
\hline
\textbf{Dynamic-R}  &99.41{\scriptsize ($\pm$ 0.08)} &92.12{\scriptsize ($\pm$ 0.29)}  &91.32{\scriptsize ($\pm$ 0.19)}   &74.64{\scriptsize ($\pm$ 1.02)} \\
\hline
\textbf{Rolled-R}      &99.29{\scriptsize ($\pm$ 0.09)} &91.53{\scriptsize ($\pm$ 0.22)}  &90.75{\scriptsize ($\pm$ 0.52)}   &74.26{\scriptsize ($\pm$ 0.94)}  \\
\hline
\textbf{Trainable-R} &99.55{\scriptsize ($\pm$ 0.04)} &92.58{\scriptsize ($\pm$ 0.10)}  &92.37{\scriptsize ($\pm$ 0.29)}  &76.43{\scriptsize ($\pm$ 1.11)}  \\
\hline
\textbf{No-R}           &99.54{\scriptsize ($\pm$ 0.04)}  &92.53{\scriptsize ($\pm$ 0.26)}  &92.15{\scriptsize ($\pm$ 0.29)}  &76.28{\scriptsize ($\pm$ 0.39)} \\
\hline
\end{tabular}
\end{center}
\caption{The performance of CapsNets with different routing procedures on different standard datasets is shown, where the standard (untransformed) test datasets are used. We can observe that the routing procedures do not improve performance.}
\label{tab:perf}
\end{table*}

\subsection{Effectiveness of the Dynamic Routing}
\label{exp:effect}
In Section \ref{sec:effect}, we show that the routing mechanism can be learned implicitly in CapsNets without routing procedure. Our experiments in this section aim to investigate if the advantages of CapsNets disappear when trained with no routing. We consider the following routing procedures in our training routines:
\begin{enumerate}
    \item \textbf{Dynamic-R}: with standard dynamic routing in \cite{sabour2017dynamic};
	\item \textbf{Rolled-R}: with a rolled routing procedure by treating coupling coefficients as constants during Gradient Backpropagation, as analyzed in Section \ref{sec:rolled};
	\item \textbf{Trainable-R}: one-step routing with trainable coupling coefficients, as in \cite{chen2018generalized};
	\item \textbf{No-R}: without routing procedure, which is equivalent to the uniform routing in \cite{Paik2019CapsuleNN,anonymous2020capsule}.
\end{enumerate}
We train CapsNets with different routing procedures described above on four standard datasets, namely, MNIST \cite{lecun1998gradient}, FMNIST \cite{xiao2017fashion}, SVHN \cite{netzer2011reading} and CIFAR10 \cite{krizhevsky2009learning}. The performance is reported in Table \ref{tab:perf}. 

Given the performance variance for each model, the performance between different models is relatively small. The reason behind this is that coupling coefficients can be learned in transformation matrices implicitly, and all the models possess a similar transformation process. The models trained with \textbf{No-R} do not prevent the learning of coupling coefficients. We can also observe that the models with \textbf{Trainable-R} or \textbf{No-R} show a slightly better performance than the other two. To our understanding, the reason is that they do not suffer the polarization problem of coupling coefficients \cite{li2018neural}.

From this experiment, we can only conclude that the routing procedure does not contribute to the generalization ability of CapsNets. In work \cite{sabour2017dynamic}, CapsNets show many superior properties over CNNs, besides the classification performance. In the following, we analyze the properties of CapsNets with \textbf{No-R} and compare them with CapsNets with \textbf{Dynamic-R}.

\subsubsection{On learned Representations of Capsules}
When training CapsNets, the original input is reconstructed from the activity vector (i.e., instantiation parameters) of a high-level capsule. The reconstruction is treated as a regularization technique.  In CapsNets with \textbf{Dynamic-R} \cite{sabour2017dynamic}, the dimensions of the activity vector learn how to span the space containing large variations. To check such property of CapsNets with \textbf{No-R}, following \cite{sabour2017dynamic}, we feed a perturbed activity vector of the ground-truth class to decoder network. 

\begin{figure}[h]
  \centering
  \includegraphics[width=\linewidth]{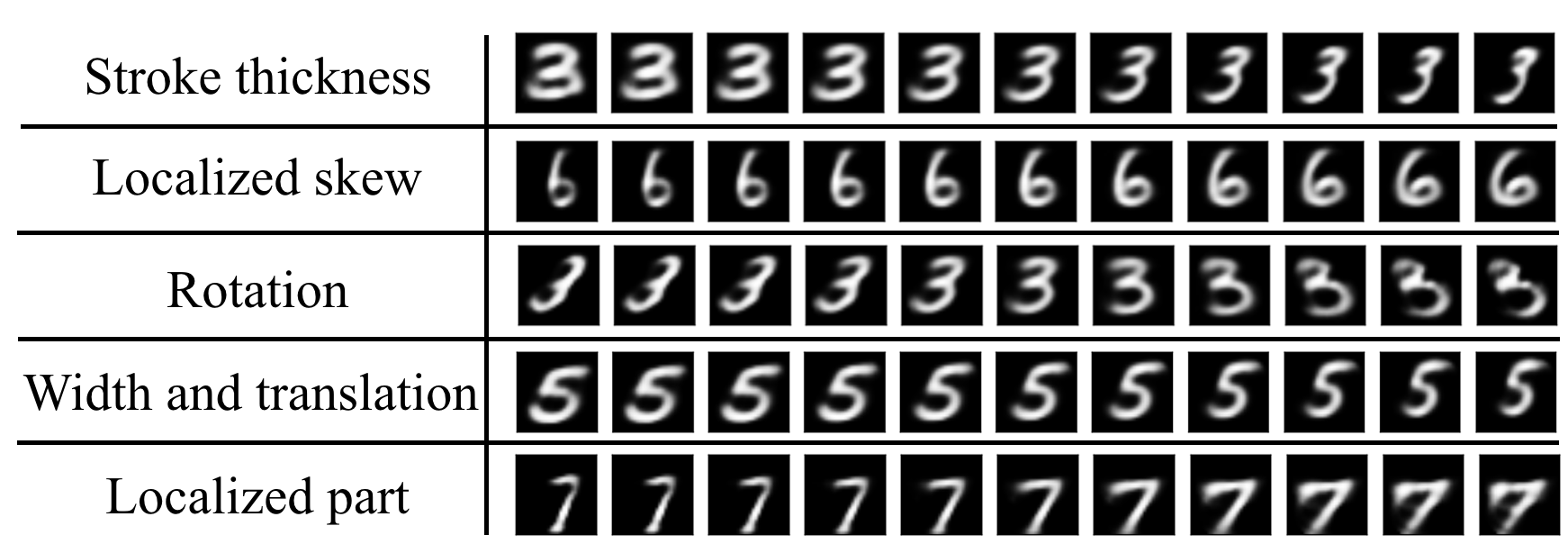}
  \caption{Disentangled Individual Dimensions of Capsules: By perturbing one dimension of an activity vector, the variations of an input image are reconstructed.}
  \label{fig:disent}
\end{figure}

\begin{figure}[h]
  \centering
  \includegraphics[width=0.85\linewidth]{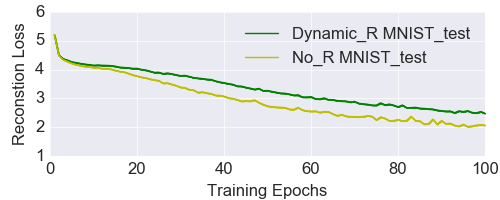}
 \caption{The average reconstruction loss of CapsNets with \textbf{Dynamic-R} and \textbf{No-R} on the test dataset is shown in each epoch of the training process.}
  \label{fig:recons_loss}
\end{figure}

The perturbation of the dimensions can also cause variations of the reconstructed input. We show some examples in Figure \ref{fig:disent}. The variations include stroke thickness, width, translation, rotation, and various combinations.  In Figure \ref{fig:recons_loss}, we also visualize the reconstruction loss of the models with \textbf{Dynamic-R} and the ones with \textbf{No-R}. The CapsNets with \textbf{No-R} show even less reconstruction error and can reconstruct inputs better.

\subsubsection{Parallel Attention Mechanism between Capsules}
Dynamic routing can be viewed as a parallel attention mechanism, in which each high-level capsule attends to some active low-level capsules and ignores others. The parallel attention mechanism allows the model to recognize multiple objects in the image even if objects overlap \cite{sabour2017dynamic}. The superiority of the parallel attention mechanism can be shown on the classification task on MultiMNIST dataset \cite{hinton2000learning,sabour2017dynamic}. Each image in this dataset contains two highly overlapping digits. CapsNet with dynamic routing procedure shows high performance on this task.

In this experiment, we show that the parallel attention mechanism between capsules can be learned implicitly, even without the routing mechanism. Following the experimental setting in \cite{sabour2017dynamic}, we train a CapsNet with \textbf{No-R} on the same classification task of classifying highly overlapping digits. The model \textbf{No-R} achieves 95,49\% accuracy on the test set, while the one with \textbf{Dynamic-R} achieves 95\% accuracy. The removal of the routing procedure does not make the parallel attention mechanism of CapsNets disappear.

\begin{figure*}
    \centering
    \begin{subfigure}[b]{0.48\textwidth}
        \includegraphics[width=\textwidth]{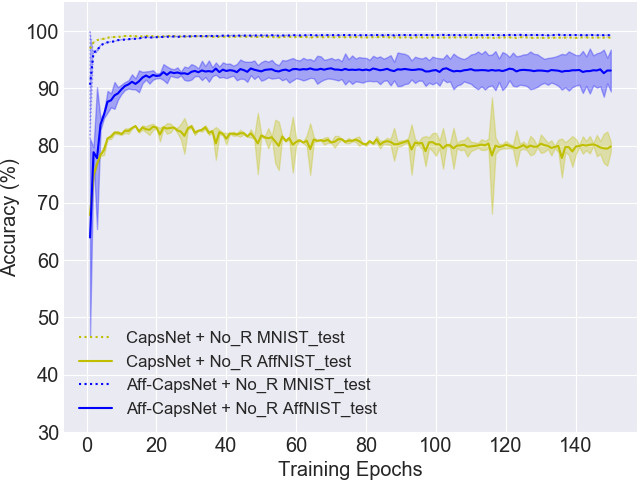}
        \caption{Without a routing procedure: the test accuracy of CapsNets and Aff-Capsnets on on the expanded MNIST test set and the AffNIST test set.}
        \label{fig:withNoR}
    \end{subfigure} \hspace{0.5cm}
    \begin{subfigure}[b]{0.48\textwidth}
        \includegraphics[width=\textwidth]{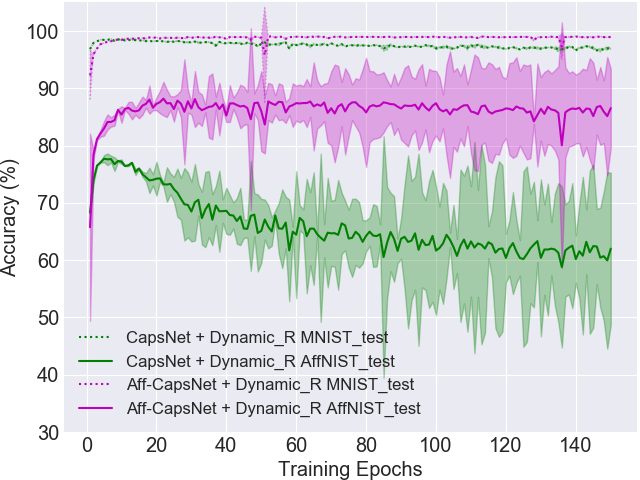}
        \caption{With the dynamic routing: the test accuracy of CapsNets and Aff-Capsnets on the expanded MNIST test set and the AffNIST test set.}
        \label{fig:withDR}
    \end{subfigure}
    \caption{For both cases (with or without routing procedure), Aff-CapsNets clearly outperform CapsNets on the AffNIST test dataset.}
    \label{fig:affcaps}
\end{figure*}

\subsubsection{Robustness to Affine Transformation}
CapsNets are also known for their robustness to affine transformation. It is important to check whether the removal of the routing procedure affects the affine robustness. We conduct experiments on a standard benchmark task. Following \cite{sabour2017dynamic}, we train CapsNets with or without routing procedure on the MNIST training dataset and test them on the affNIST dataset. The images in the MNIST training dataset are placed randomly on a black ground of $40 \times 40$ pixels to match the size of images in affNIST dataset. The CNN baseline is set the same as in \cite{sabour2017dynamic}.

It is hard to decide if one model is better at generalizing to novel affine transformations than another one when they achieved different accuracy on untransformed examples. To eliminate this confounding factor, we stopped training the models when they achieve similar performance, following \cite{sabour2017dynamic}. The performance is shown in Table \ref{tab:affNoR}. Without routing procedure, the CapsNets show even better affine robustness. 

In summary, our experiments show that the dynamic routing procedure contributes neither to the generalization ability nor to the affine robustness. Due to the high affine robustness of CapsNet cannot be attributed to the routing procedure: Instead, it is the inductive bias (architecture) of CapsNets that contributes to the affine robustness.

\subsection{Affine Robustness of Aff-CapsNets}
In Section \ref{sec:aff}, we proposed Aff-CapsNets that are more robust to the novel affine transformations of inputs. In this experiment, we train Aff-CapsNets with \textbf{Dynamic-R} and \textbf{No-R} respectively. As a comparison, we also train CapsNets with or without dynamic routing correspondingly.

We visualize the test accuracy on the expanded MNIST test set and the AffNIST test set. The performance is shown in Figure \ref{fig:affcaps}. The lines show the averaged values, while the colored areas around the lines describe the variances caused by different seeds. Figure \ref{fig:withNoR} shows the accuracy of models trained without a routing procedure. We can observe that the Aff-CapsNets constantly shows better accuracy than CapsNets on AffNIST. To a great extent, our Aff-CapsNets covers the performance gap between the test accuracy on untransformed examples and that on transformed ones. 

\begin{table}
\begin{center}
\begin{tabular}{rP{2.4cm}P{2.4cm}}
\toprule
Models  & Test on MNIST &  Test on AffNIST  \\
\midrule
CNN \cite{sabour2017dynamic}  & 99.22\%  & 66\% \\
\textbf{Dynamic-R} \cite{sabour2017dynamic}  & 99.23\% &  79\%\\
\textbf{No-R}  & 99.22\% &   81.81\% \\
\bottomrule
\end{tabular}
\end{center}
\caption{The performance on the expanded MNIST test set and the AffNIST test set.}
\label{tab:affNoR}
\end{table}

In addition, the Aff-CapsNet architecture is still effective, even when the dynamic routing is applied in training (see Figure \ref{fig:withDR}). We can also observe that the CapsNets with dynamic routing overfit to the current viewpoints. With the training process going on, the coupling coefficients are polarized (become close to 0 or 1) \cite{li2018neural}. The polarization of the coupling coefficient causes the overfitting. Furthermore, the training with dynamic routing is more unstable than without routing. The variance of model test performance in  Figure \ref{fig:withDR} is much bigger than the ones in  Figure \ref{fig:withNoR}.

We now compare our model with previous work. In Table \ref{tab:affNoRSOTA}, we list the performance of CNN variants and CapsNet variants on this task. Without training on AffNIST dataset, our Aff-CapsNets achieve state-of-the-art performance on AffNIST test dataset. This experiment shows that the proposed model is robust to input affine transformation.

\section{Discussion}
\textbf{The difference between the regular CNNs, Aff-Capsnet and CapsNets:} Each neuron in the convolutional layer is connected only to a local spatial region in the input. However, each element in a capsule layer (with or without dynamic routing) is connected to all elements of all input capsules. By considering global information, the features extracted by the capsule layer might be more useful for some tasks, e.g., affine-transformed image classification or semantic image segmentation.

What is the difference between a fully connected (FC) layer and the capsule layer without dynamic routing? In an FC layer, each neuron is also connected to all neurons of the preceding layer. Compared with FC layers, convolutional layers show inductive biases, which are Local Connection and Parameter Sharing. Similarly, capsule layers might show a new inductive bias, namely, a new way to combine activations of the preceding layer.

\begin{figure}[t]
\begin{center}
   \includegraphics[scale=0.368]{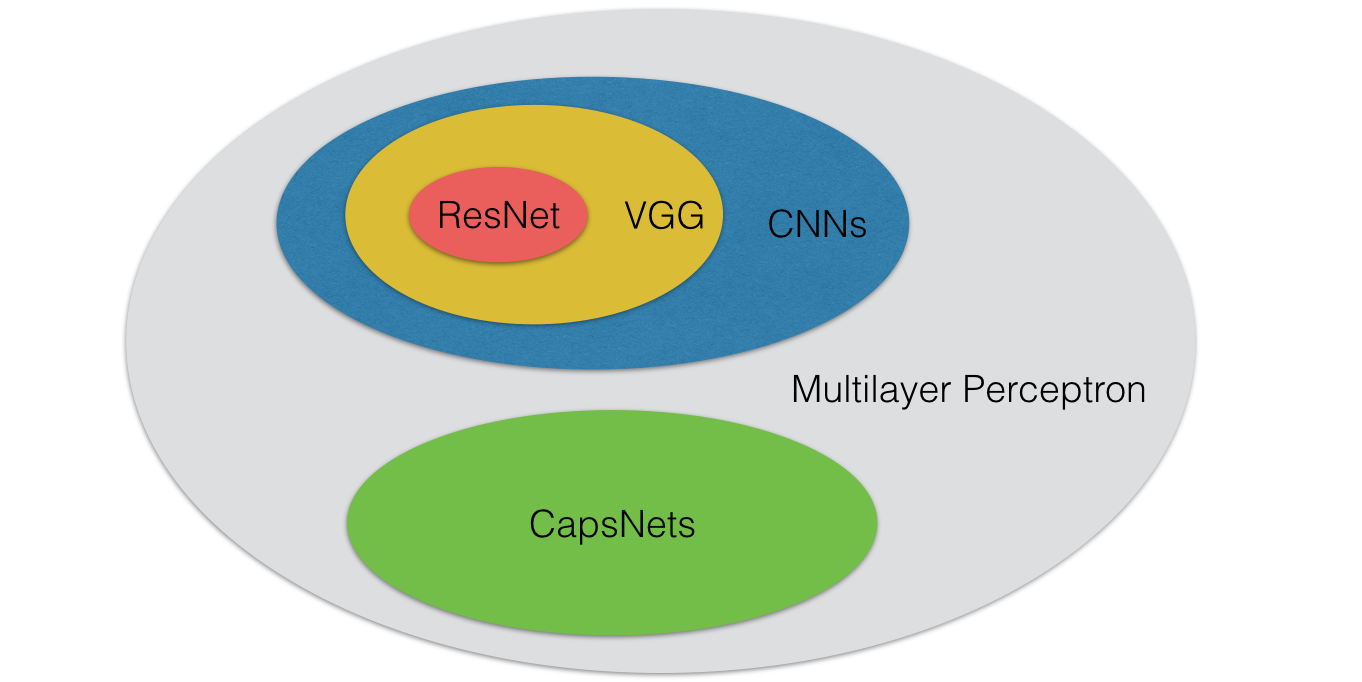}
\end{center}
   \caption{The relationship between different CNN architectures and Capsule Network architectures.}
\label{fig:sum}
\end{figure}

The relationship between CapsNet architectures and CNN architectures is illustrated in Figure \ref{fig:sum}. CapsNets might be considered as new architectures parallel to CNNs. In the past years, our community has focused on exploring CNN architectures manually or automatically. The figure illustrates that there is "space" outside of the CNN paradigm: CapsNets, or even other unexplored options.

\begin{table}[t]
\begin{center}
\small
\begin{tabular}{P{2.7cm}P{1.4cm}P{1.1cm}P{1.6cm}}
\toprule
Models  & Trained on AffNIST? &  MNIST &  AffNIST  \\
\midrule
Marginal. CNN \cite{zhao122017marginalized} &Yes  & 97.82\%& 86.79\%\\
TransRA CNN\cite{Asano2018} &Yes  & 99.25 \%& 87.57\%\\
BCN \cite{chang2018broadcasting} &Mix* & 97.5\%& 91.60\%\\
\midrule
CNN \cite{sabour2017dynamic} & No  & 99.22\%  & 66\% \\
Dynamic-R \cite{sabour2017dynamic} & No  & 99.23\% &  79\%\\
GE-CAPS \cite{lenssen2018group}& No  & - & 89.10\% \\
SPARSECAPS \cite{rawlinson2018sparse} & No  & 99\% & 90.12\%\\
Aff-CapsNet + \textbf{No-R}& No  & 99.23\% & \textbf{93.21{\tiny($\pm$0.65)}\%} \\
\bottomrule
\end{tabular}
\end{center}
\caption{Comparison to state-of-the-art performance on the benchmark task.}
\label{tab:affNoRSOTA}
\end{table}

\textbf{Going Deeper with CapsNets:} One way to make CapsNets deep is to integrate advanced techniques of training CNNs into CapsNets. The integration of skip connections \cite{he2016deep,rajasegaran2019deepcaps} and dense connections \cite{huang2017densely,phaye2018multi} have been proven to be successful. Instead of blindly integrating more advanced techniques from CNN into CapsNets, it might be more promising to investigate more into the effective components in CapsNets. Our investigation reveals that the dynamic routing procedure contributes neither to the generalization ability nor to the affine robustness of CapsNets. Such conclusion is helpful for training CapsNets on large scale datasets, e.g., the ImageNet 1K dataset \cite{deng2009imagenet}.

\textbf{Application of CapsNets to Computer Vision Tasks} Besides the object recognition task, CapsNets are also applied to many other computer vision tasks, for examples, object segmentation \cite{lalonde2018capsules}, image generation models \cite{jaiswal2018capsulegan,saqur2019capsgan}, and adversarial defense \cite{hinton2018matrix}. It is not clear whether routing procedures are necessary for these tasks. If routing is not required here as well, the architectures of CapsuleNets can be integrated into these vision tasks with much less effort.

\textbf{The Necessity of the Routing Procedure in CapsNets} \cite{sabour2017dynamic} demonstrated many advantages of CapsNets with dynamic routing over CNNs. However, our investigation shows that all the advantages do not disappear when the routing procedure is removed. Our paper does not claim that routing does not have any benefits but rather poses the question to the community: \textit{What is the routing procedure really good for?} If the routing procedure is not necessary for a given task, CapsNets have the chance of becoming an easier-to-use building block.

\section{Conclusion}
We revisit the dynamic routing procedure of CapsNets. Our numerical analysis and extensive experiments show that neither the generalization ability nor the affine robustness of CapsNets is reduced by removing the dynamic routing procedure. This insight guided us to focus on the CapsNet architecture, instead of various routing procedures, to improve the affine robustness. After exploring the limitation of the CapsNet architecture, we propose Aff-CapsNets, which improves affine robustness significantly using fewer parameters.

Since this work mainly focused on the robustness to affine transformation, we investigate the standard CapsNets with dynamic routings. Other beneficial properties have also been shown in improved CapsNets, like adversarial robustness and viewpoint invariance. Further analysis of these properties will be addressed in future work.

\newpage
\bibliographystyle{ieee_fullname}
\bibliography{egbib}

\begin{thebibliography}{10}\itemsep=-1pt

\bibitem{Asano2018}
Shuhei Asano.
\newblock {\em Proposal of transformation robust attentive convolutional neural
  network}.
\newblock PhD thesis, Waseda University, 2018.

\bibitem{bahadori2018spectral}
Mohammad~Taha Bahadori.
\newblock Spectral capsule networks.
\newblock In {\em ICML Workshop}, 2018.

\bibitem{chang2018broadcasting}
Simyung Chang, John Yang, SeongUk Park, and Nojun Kwak.
\newblock Broadcasting convolutional network for visual relational reasoning.
\newblock In {\em ECCV}, pages 754--769, 2018.

\bibitem{chen2018generalized}
Zhenhua Chen and David Crandall.
\newblock Generalized capsule networks with trainable routing procedure.
\newblock In {\em ICML Workshop}, 2019.

\bibitem{anonymous2020capsule}
Zhenhua Chen, Xiwen Li, Chuhua Wang, and David Crandall.
\newblock Capsule networks without routing procedures.
\newblock In {\em ICLR open review submissions}, 2020.

\bibitem{Choi2019AttnCaps}
Jaewoong Choi, Hyun Seo, Suii Im, and Myungjoo Kang.
\newblock Attention routing between capsules.
\newblock In {\em Proceedings of the IEEE International Conference on Computer
  Vision Workshops}, pages 0--0, 2019.

\bibitem{deng2009imagenet}
Jia Deng, Wei Dong, Richard Socher, Li-Jia Li, Kai Li, and Li Fei-Fei.
\newblock Imagenet: A large-scale hierarchical image database.
\newblock In {\em CVPR}, pages 248--255. Ieee, 2009.

\bibitem{he2016deep}
Kaiming He, Xiangyu Zhang, Shaoqing Ren, and Jian Sun.
\newblock Deep residual learning for image recognition.
\newblock In {\em CVPR}, pages 770--778, 2016.

\bibitem{hinton2000learning}
Geoffrey~E Hinton, Zoubin Ghahramani, and Yee~Whye Teh.
\newblock Learning to parse images.
\newblock In {\em Advances in neural information processing systems}, pages
  463--469, 2000.

\bibitem{hinton2018matrix}
Geoffrey~E Hinton, Sara Sabour, and Nicholas Frosst.
\newblock Matrix capsules with em routing.
\newblock In {\em ICLR}, 2018.

\bibitem{huang2017densely}
Gao Huang, Zhuang Liu, Laurens Van Der~Maaten, and Kilian~Q Weinberger.
\newblock Densely connected convolutional networks.
\newblock In {\em CVPR}, pages 4700--4708, 2017.

\bibitem{jaiswal2018capsulegan}
Ayush Jaiswal, Wael AbdAlmageed, Yue Wu, and Premkumar Natarajan.
\newblock Capsulegan: Generative adversarial capsule network.
\newblock In {\em ECCV}, pages 0--0, 2018.

\bibitem{krizhevsky2009learning}
Alex Krizhevsky et~al.
\newblock Learning multiple layers of features from tiny images.
\newblock 2009.

\bibitem{lalonde2018capsules}
Rodney LaLonde and Ulas Bagci.
\newblock Capsules for object segmentation.
\newblock In {\em International Conference on Medical Imaging with Deep
  Learning}, 2018.

\bibitem{lecun1998gradient}
Yann LeCun, L{\'e}on Bottou, Yoshua Bengio, and Patrick Haffner.
\newblock Gradient-based learning applied to document recognition.
\newblock {\em Proceedings of the IEEE}, 86(11):2278--2324, 1998.

\bibitem{lenssen2018group}
Jan~Eric Lenssen, Matthias Fey, and Pascal Libuschewski.
\newblock Group equivariant capsule networks.
\newblock In {\em Advances in Neural Information Processing Systems}, pages
  8844--8853, 2018.

\bibitem{li2018neural}
Hongyang Li, Xiaoyang Guo, Bo DaiWanli~Ouyang, and Xiaogang Wang.
\newblock Neural network encapsulation.
\newblock In {\em ECCV}, pages 252--267, 2018.

\bibitem{netzer2011reading}
Yuval Netzer, Tao Wang, Adam Coates, Alessandro Bissacco, Bo Wu, and Andrew~Y
  Ng.
\newblock Reading digits in natural images with unsupervised feature learning.
\newblock 2011.

\bibitem{Paik2019CapsuleNN}
Inyoung Paik, Taeyeong Kwak, and Injung Kim.
\newblock Capsule networks need an improved routing algorithm.
\newblock {\em ArXiv}, abs/1907.13327, 2019.

\bibitem{phaye2018multi}
Sai Samarth~R Phaye, Apoorva Sikka, Abhinav Dhall, and Deepti~R Bathula.
\newblock Multi-level dense capsule networks.
\newblock In {\em Asian Conference on Computer Vision}, pages 577--592.
  Springer, 2018.

\bibitem{rajasegaran2019deepcaps}
Jathushan Rajasegaran, Vinoj Jayasundara, Sandaru Jayasekara, Hirunima
  Jayasekara, Suranga Seneviratne, and Ranga Rodrigo.
\newblock Deepcaps: Going deeper with capsule networks.
\newblock In {\em CVPR}, pages 10725--10733, 2019.

\bibitem{rawlinson2018sparse}
David Rawlinson, Abdelrahman Ahmed, and Gideon Kowadlo.
\newblock Sparse unsupervised capsules generalize better.
\newblock {\em arXiv preprint arXiv:1804.06094}, 2018.

\bibitem{sabour2017dynamic}
Sara Sabour, Nicholas Frosst, and Geoffrey~E Hinton.
\newblock Dynamic routing between capsules.
\newblock In {\em Advances in neural information processing systems}, pages
  3856--3866, 2017.

\bibitem{saqur2019capsgan}
Raeid Saqur and Sal Vivona.
\newblock Capsgan: Using dynamic routing for generative adversarial networks.
\newblock In {\em Science and Information Conference}, pages 511--525.
  Springer, 2019.

\bibitem{wang2018optimization}
Dilin Wang and Qiang Liu.
\newblock An optimization view on dynamic routing between capsules.
\newblock In {\em ICLR Worksop}, 2018.

\bibitem{xiao2017fashion}
Han Xiao, Kashif Rasul, and Roland Vollgraf.
\newblock Fashion-mnist: a novel image dataset for benchmarking machine
  learning algorithms.
\newblock {\em arXiv preprint arXiv:1708.07747}, 2017.

\bibitem{zhang2018fast}
Suofei Zhang, Quan Zhou, and Xiaofu Wu.
\newblock Fast dynamic routing based on weighted kernel density estimation.
\newblock In {\em International Symposium on Artificial Intelligence and
  Robotics}, pages 301--309. Springer, 2018.

\bibitem{zhao122017marginalized}
Jian Zhao, Jianshu Li, Fang Zhao, Xuecheng Nie, Yunpeng Chen, Shuicheng Yan,
  and Jiashi Feng.
\newblock Marginalized cnn: Learning deep invariant representations.
\newblock In {\em BMVC}, 2017.

\end{thebibliography}

\end{document}